# Classification of Smartphone Users Using Internet Traffic

Andrey Finkelstein, Ron Biton, Rami Puzis, Asaf Shabtai
Department of Software and Information Systems Engineering
Ben-Gurion University of the Negev, Beer-Sheva, Israel

## ABSTRACT

Today, smartphone devices are owned by a large portion of the population and have become a very popular platform for accessing the Internet. Smartphones provide the user with immediate access to information and services. However, they can easily expose the user to many privacy risks. Applications that are installed on the device and entities with access to the device's Internet traffic can reveal private information about the smartphone user and steal sensitive content stored on the device or transmitted by the device over the Internet. In this paper, we present a method to reveal various demographics and technical computer skills of smartphone users by their Internet traffic records, using machine learning classification models. We implement and evaluate the method on real life data of smartphone users and show that smartphone users can be classified by their gender, smoking habits, software programming experience, and other characteristics.

## INTRODUCTION

In recent years, the number of smartphone users has rapidly increased. According to a report published by Smart Insights[1], the number of smartphone users grew from 400 million users in 2007, to more than 1,800 million in 2015. In addition, the report claims that at the end of 2015, 97% of adults, aged 18 to 34, in the US were mobile device users. The mobility and capabilities of smartphones make them a very popular platform for Internet usage. According to [1], approximately two thirds of the adult population (ages 16 and over) in the UK use smartphones to go online, and the number increases to 90% among adults aged 16 to 34.

The various functionalities of smartphones make them very useful devices, however these capabilities also pose a great privacy risk to smartphone users [2]. In many cases, smartphone users store sensitive information such as private photos and passwords on their devices. Moreover, smartphones give applications access to sensors such as GPS, gyroscope, and accelerometer. These, and other sensors, can be used to reveal information about the user, including activity recognition [3] and demographic properties (e.g., gender) [4], by malicious applications installed on the device.

However, the privacy risks smartphone users are exposed to are not limited to device applications. The Internet exposes smartphone users to many other entities that may violate user privacy. Public Wi-Fi networks, ISP providers, VPN (virtual private network) services, and proxy servers are examples of entities that have access to the Internet traffic of smartphone users. This traffic may contain sensitive information transmitted in plain text (e.g., HTTP forms). However, recent studies show that private information regarding the user may be extracted from Internet traffic using machine learning techniques. In [5], the authors extract statistical and application and category-based features from the traffic and use location properties of the hotspot to show how public Wi-Fi can reveal the gender and education of its users. In [6], a scenario where remote entities with access to the user's smartphone Internet (e.g., VPN services) can use it to identify the type of venue (home, organization, hangout, and waiting place) the user is located at.

In this paper, we present a method for classifying smartphone users by various demographic properties and computer technical skills. The method analyzes and aggregates smartphones' Internet traffic records to extract features that represent the smartphone user. The feature extraction process uses feature extraction techniques that were introduced in [5] and [7], which are enriched with additional new features defined in this study. By applying a supervised machine learning approach, we were able to classify smartphone users by 10 different properties including their gender, age group, and education. The method was demonstrated and evaluated on real data (network traffic) of 143 smartphone users collected during 2014 and 2015; for example, we were able to classify the users by their gender and software programing experience with an *accuracy* of 83.9% and 77.8%, respectively.

## METHOD

Mobile Internet traffic datasets are not publicly available due to their sensitive nature in terms of privacy. Therefore, first we had to collect such data from smartphone users by conducting an experiment. In addition, at the start of the experiment, the users were asked to complete a questionnaire to tell us about themselves and their technical computer skills. After the data was collected, the traffic records were processed and aggregated. We extracted four main groups of features, as well as the set of demographic and technical computer skills of the subjects that were used as labels. Finally, a supervised machine learning framework was applied to train and evaluate classification models on the data obtained during the experiment.

### Experiment Set-Up

143 students with Android devices participated in the experiment. We divided the experiment into four parts. The first two parts were conducted during 2014, and consisted of a month-long data collection involving 17 subjects, and a slightly longer (two months) data collection process involving 61 subjects. The other two parts of the experiment took place during 2015 and involved a total of 65 subjects.

---

[1] http://www.smartinsights.com/mobile-marketing/mobile-marketing-analytics/mobile-marketing-statistics/

At the start of the experiment all of the subjects had to complete a questionnaire provide, in order to provide their demographical characteristics (e.g., age and gender) and technical skills regarding computers and programming (e.g., whether the subject ever wrote a short code or formatted a computer). Figure 1 presents the distribution of the subjects for several demographic and technological properties.

The subjects were requested to install a VPN (virtual private network) client on their devices called OpenVPN Connect which is available on the Google Play application market. The VPN client was used to redirect the subjects' Internet traffic through the experiment's dedicated VPN server where the traffic was recorded and stored until the end of the experiment. A unique configuration was set to each subject's device, so the subject's traffic could be distinguished from other subjects' traffic on the server side.

The subjects were requested to stay connected to the VPN server continuously during the entire experimental period. However, disconnection events occurred often, due to many reasons such as poor network signal, change of networks (e.g., user switched from Wi-Fi to 3G connection), issues with the Android VPN API, and subject initiated disconnections.

Note that the experiment was approved by the university ethical committee.

**Data Processing**

Once the data collection experiment was completed, the data was transferred securely to an analytical server for data processing as follows.

First, the traffic records were aggregated into sessions in a manner similar to the method introduced in [7]. Sessions were defined as one of the following: a TCP session (SYN to FYN) or a UDP request and its response. Then, for every session we extracted features from four different feature groups: *statistical features*, *application layer features*, *domain features*, and *deep packet inspection features*:

**Statistical Features** – A subset of the feature set that was introduced in [7]. Mainly consisting of the traffic volume features: transmitted and received packets' size statistics (max, min, mean, median, and variance), the number of bytes within a session (total, transmitted, and received), and the ratio between transmitted and received traffic.

We decided not to use features that represent networks' quality of service, which were introduced in [7] (e.g., the number of retransmitted packets or the interval time between packets), since the focus of the study is to classify the user and not the network.

**Application Layer Features** – We focused on the two most common protocols in the data: HTTPS and HTTP. In HTTPS sessions we examined the SSL/TLS version used to determine the connection's security, and we checked that the SSL certificate was not expired or self-signed. From HTTP sessions we extracted the number of cookies the client provided and the Content-Type field, similar to [7]. In addition, we parsed the User-Agent string to extract the OS version of the device.

Another piece of information that we extracted from these protocols was the domain name (HTTP hostname and SSL server name). The domain names were used to ex-tract the *domain features*.

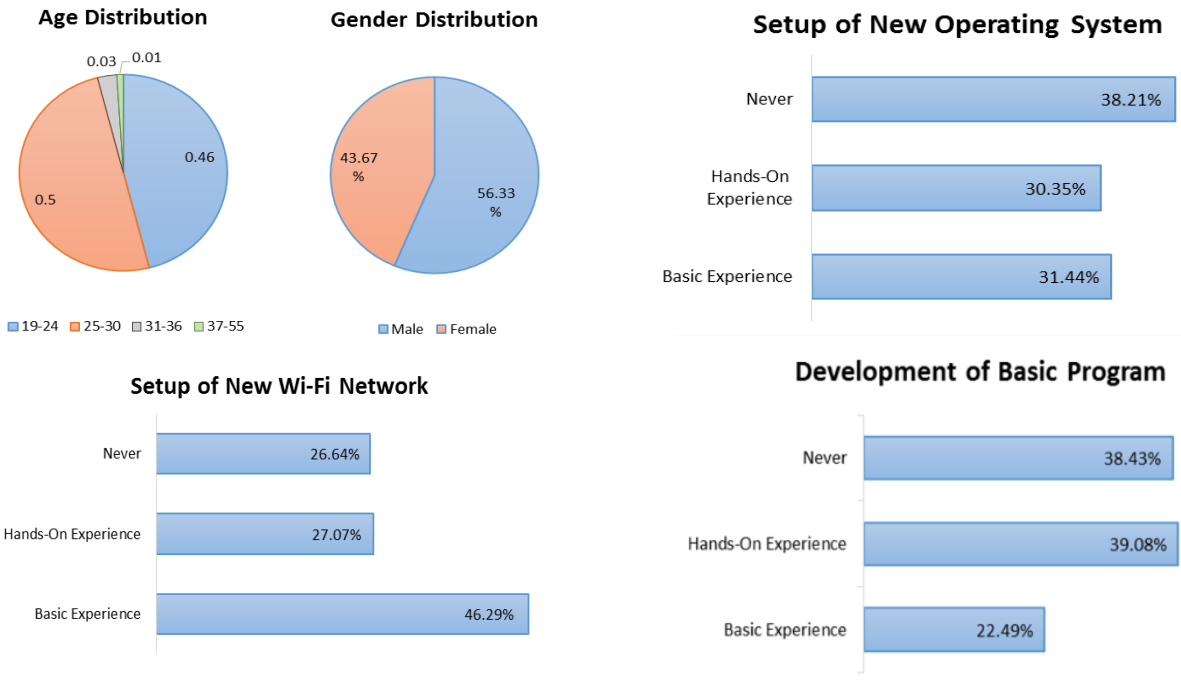

**Fig. 1. Subjects' demographics and technical computer skills**

**Domain Features** – The domain names that were extracted from the application layer headers can provide various information about the mobile user. To extract such information we used the following third party services and databases: VirusTotal, alexa rank, WoT (web of trust), BitDefender category, and "urlblacklist.com.

For every session with a domain name available, we extracted the "alexa" popularity score. We used "WoT" to determine domains' security scores (good site, trustworthiness, and child safety) and security categories (scam, spam, malware or viruses, privacy risks, and phishing). In addition, a general category such as social network or education (32 possible values) was derived by combining and aggregating the BitDefender and UrlBlacklist.com domain categories.

**Deep Packet Inspection Features** – Deep packet inspection is a complicated process where the content of the packet is analyzed. It can provide meaningful information about the subject, however it takes significant resources and effort to extract this information. We counted the number of HTTP forms, the presence of email addresses, usernames, and password fields in these forms, the number of downloaded files and their types.

The deep packet inspection process included decoding GZIP encoded traffic and parsing JSON and XML files which are very common in HTTP traffic.

A single session may not contain enough information for reliable profiling of users. Thus, all of the sessions that were associated with a subject were aggregated to a single instance. The aggregation process was performed as follows. For every numerical session feature, we calculated the average, median, minimal, and maximal values across all of the sessions associated with the subject. For nominal session features (e.g., domain category features), we created numerical subject features that represent the categorical value's incidence in the subject's sessions. For example, if 50 sessions were associated with a subject, of which 30 were from the "search" category and 20 were categorized as "news," the values of the "search" and "news" features for the subject were 0.6 and 0.4, respectively, and the values of the other domain category features were all 0.

In addition, for each subject we extracted the ratios of traffic volumes between the most popular ports in the experiment's traffic (TCP 80 – HTTP, TCP 443 – HTTPS, and TCP 5228 – Google Play store).

To form the demographic and technical computer skills dataset, the questions from the questionnaire were used as labels for the subjects. Table 1 presents the labels that were extracted for each subject and the values of the labels in the dataset.

**Table 1. User classification labels and classes**

| *Labels* | *Classes* |
|---|---|
| Age Group | 18-24, 25-30, 31+ |
| Gender | Male, Female |
| Education | Higher Education, High-School |
| Faculty | Natural Sciences, Humanities, Engineering |
| Smokes | Yes, No |
| Setup new OS | Never, Basic Experience, Hands on Experience |
| **Setup** a Wi-Fi Network | Never, Basic Experience, Hands-On Experience |
| Wrote a Program | Never, Basic Experience, Hands-On Experience |
| Formatted a Computer | Never, Basic Experience, Hands-On Experience |
| Built a Website | Never, Basic Experience, Hands-On Experience |

## EVALUATION

In order to classify the demographics and technical computer skills of subjects from their mobile Internet traffic, a supervised machine learning approach was chosen.

We used the scikit-learn machine learning python packages to train machine learning classification models. For every label, we trained and evaluated multiple machine learning models that differ by the classification algorithm and number of features that were used to train the model. The machine learning models were based on the random forest (RF) and extra trees (ET) ensemble classification algorithms. Feature selection was performed using the K-best by Anova F-Value algorithm (for K values of 30, 50, 80, and 120).

To evaluate the quality of the classification models, we used the leave-one-out cross-validation method. In this method the entire dataset except for a single instance is used for training, and the excluded instance is used for testing the model. This process is repeated for every instance in the dataset, and the performance measures are calculated on the classification of all of the instances.

The evaluation metrics we present are the accuracy score, and the weighted AUC (WAUC), weighted precision, and weighted recall measures. In addition, we present the F1 score which combines the precision and recall metrics. Table 2 presents the evaluation results of the classification models that yielded the best F1 score for each label. The results show that it is possible to classify smartphone users by their demographic characteristics and technical computer skills.

To better understand how the Internet traffic reveals information about the smartphone user, we analyze the importance of the features in the different classification models. The importance of a feature for a model was defined as the feature's average importance in the decision trees which are part of the ensemble classification model.

Table 2. Smartphone users' classification results.

| Label | Algo. | Features | Accuracy | WAUC | W-Precision | W-Recall | F1 |
|---|---|---|---|---|---|---|---|
| Gender | RF | 100 | 0.839 | 0.891 | 0.851 | 0.793 | 0.821 |
| Education | ET | 100 | 0.839 | 0.845 | 0.911 | 0.688 | 0.784 |
| Wrote a Program | ET | 80 | 0.778 | 0.756 | 0.770 | 0.741 | 0.755 |
| Smokes | RF | 100 | 0.862 | 0.896 | 0.776 | 0.679 | 0.724 |
| Setup a Wi-Fi Network | ET | 80 | 0.710 | 0.797 | 0.800 | 0.654 | 0.720 |
| Age Group | ET | 120 | 0.700 | 0.779 | 0.699 | 0.696 | 0.698 |
| Formatted a Computer | ET | 30 | 0.677 | 0.854 | 0.761 | 0.640 | 0.696 |
| Setup new OS | ET | 100 | 0.677 | 0.845 | 0.678 | 0.676 | 0.677 |
| Faculty | RF | 100 | 0.600 | 0.795 | 0.607 | 0.590 | 0.599 |
| Built a Website | ET | 80 | 0.581 | 0.628 | 0.661 | 0.469 | 0.549 |

The importance of a feature in a decision tree was derived by evaluating the ability of the feature to distinguish a specific class and the depth at which the feature appears in the tree (features close to the root affect more samples). For each label, the top five features (based on their importance) were extracted from the model that presented the best WAUC results. Table 3 contains the number of features that were selected as top five features by category. It can be seen that the majority of the important features belong to the domain feature category (78%), while the other categories have nearly the same amount of representation in this table.

Table 3. Number of important features by feature category

| Feature Category | # of Important Features |
|---|---|
| Domain | 39 |
| Deep packet inspection | 4 |
| Statistical | 4 |
| Application layer | 3 |

## CONCLUSIONS

In this paper, we use Internet traffic of smartphone users to classify them by various demographic characteristics and technical computer skills. We describe the feature extraction process and machine learning training scheme and implement the method on the real life Internet traffic of 143 student subjects.

The evaluation of the method shows that Internet traffic can be used to classify smartphone users and reveal information about them to entities with access to such traffic (e.g., Wi-Fi hotspots, VPN services, ISPs). Our analysis of the classification models shows that they are heavily dependent on domain features. These features represent the popularity, security, and the categories of the websites that the users communicate with. Thus, these features can be considered private information, and revealing them may violate the privacy of the user. Moreover, machine learning models are able profile users using these features and determine the demographics and technical computer skills of smartphone users.

The solution provided by VPN services for mitigating such privacy violations was suggested by [5]. However, users must select their VPN service carefully, since the services themselves may violate users' privacy. Another solution suggested by [5] was generating dummy traffic. Dummy traffic can mislead classification models by manipulating the values of features, but generating such traffic on smartphones can cause performance issues, compromise the user's experience, drain the battery, and may cause the user to incur extra charges from the mobile carrier.

This research is limited to a relatively small number of subjects, all of whom are students who live in the same country. Thus, the experiment's sample may not adequately represent the diversity of the smartphone user population. Despite this, we believe that the classification results show that smartphone users can be classified by their Internet traffic. Moreover, a larger, more diverse dataset is likely to yield better results, due to a larger training set and greater variance across the demographic groups.

In the future, we intend to increase the sample size and diversity of the users and to classify smartphone users in other ways, including their security score.

## ACKNOWLEDGMENT

We wish to thank VirusTotal and OpenVPN for the academic licenses which were essential for conducting this research. In addition, we thank Verint Ltd. for their assistance in performing the deep packet inspection of the Internet traffic.